\documentclass[Afour,sagev,times]{sagej}

\usepackage{moreverb,url}

\usepackage[colorlinks,bookmarksopen,bookmarksnumbered,citecolor=red,urlcolor=red]{hyperref}

\newcommand\BibTeX{{\rmfamily B\kern-.05em \textsc{i\kern-.025em b}\kern-.08em
T\kern-.1667em\lower.7ex\hbox{E}\kern-.125emX}}

\def\volumeyear{2021}

\begin{document}

\runninghead{DEVCOM Army Research Laboratory}

%Artificial intelligence on games and simulators as a platform for development of military applications
\title{On games and simulators as a platform for development of artificial intelligence for command and control}

% On Games and simulators for developing artificial intelligence for military applications

%Developing artificial intelligence for military applications using games and simulators

\author{Vinicius G. Goecks\affilnum{1}, Nicholas Waytowich\affilnum{1}, Derrik E. Asher\affilnum{1}, Song Jun Park\affilnum{1}, Mark Mittrick\affilnum{1}, John Richardson\affilnum{1}, Manuel Vindiola\affilnum{1}, Anne Logie\affilnum{1}, Mark Dennison\affilnum{2}, Theron Trout\affilnum{1}, Priya Narayanan\affilnum{1}, and Alexander Kott\affilnum{1}}

\affiliation{\affilnum{1}DEVCOM Army Research Laboratory, USA\\
\affilnum{2}DEVCOM Army Research Laboratory West, USA}

\corrauth{Vinicius G. Goecks,
DEVCOM Army Research Laboratory,
Human Research and Engineering Directorate,
Aberdeen Proving Ground, MD. USA.}

\email{vinicius.goecks@gmail.com}

% ----------------------------------------------------------------------
%                            ABSTRACT 
% ----------------------------------------------------------------------

\begin{abstract}
Games and simulators can be a valuable platform to execute complex multi-agent, multiplayer, imperfect information scenarios with significant parallels to military applications: multiple participants manage resources and make decisions that command assets to secure specific areas of a map or neutralize opposing forces.
These characteristics have attracted the artificial intelligence (AI) community by supporting development of algorithms with complex benchmarks and the capability to rapidly iterate over new ideas.
The success of artificial intelligence  algorithms in real-time strategy games such as \textit{StarCraft II} have also attracted the attention of the military research community aiming to explore similar techniques in  military counterpart scenarios.
Aiming to bridge the connection between games and military applications, this work discusses past and current efforts on how games and simulators, together with the artificial intelligence algorithms, have been adapted to simulate certain aspects of military missions and how they might impact the future battlefield. This paper also investigates how advances in virtual reality and visual augmentation systems open new possibilities in human interfaces with gaming platforms and their military parallels.
\end{abstract}


\maketitle

% ----------------------------------------------------------------------
%                            INTRODUCTION 
% ----------------------------------------------------------------------
\section{Introduction}

In warfare, the ability to anticipate the consequences of the opponent's possible actions and counteractions is an essential skill \cite{caffrey2019wargaming}.
This becomes even more challenging in the future battlefield and multi-domain operations where the speed and complexity of operations are likely to exceed the cognitive abilities of a human command staff in charge of conventional, largely manual, command and control processes.  
Strategy games such as chess, poker, and Go have abstracted some of command and control (C2) concepts.
While wargames are often manual games played with physical maps, markers, and spreadsheets to compute battle outcomes \cite{perla1990art}, computer games and modern game engines such as \textit{Unreal Engine}\endnote{\textit{Unreal Engine} game engine: https://www.unrealengine.com/.} and \textit{Unity}\endnote{\textit{Unity} game engine: https://unity.com/} are capable of automating simulation of complex battle and related physics simulations.

Computer games and game engines are not only suitable to simulate military scenarios but also provide a useful test-bed for developing AI algorithms. Games such as \textit{StarCraft II}\cite{vinyals2017starcraft,sun2018tstarbots,vinyals2019grandmaster,waytowich2019grounding, waytowich2019ICML, wang2020scc,han2020tstarbot}, \textit{Dota 2} \cite{openai2019dota}, \textit{Atari}\cite{mnih2013playing,mnih2015human,hessel2018rainbow}, \textit{Go}\cite{silver2016mastering,silver2017mastering,schrittwieser2020mastering}, chess\cite{campbell2002deep,hsu2002behind}, heads-up no-limit poker\cite{brown2018superhuman} have all been used as platforms for training artificially intelligent agents.
Another capability developed by the gaming industry that is relevant to the military domain is virtual reality and visual augmentation systems, which may potentially provide commanders and staff a more intuitive and information-rich display of the battlefield.

Recently, there has been a major effort towards leveraging the capabilities of computer games and simulation engines due to their suitability for integrating AI algorithms, some examples of which we discuss shortly. The work described in this paper focuses on adapting imperfect information real-time strategy games (RTS) and their corresponding simulators, to emulate military assets in battlefield scenarios. Further, this work aims to advance the research and development efforts associated with AI algorithms for command and control. %, and present a new AI algorithm for this task.
Flexibility in these games and game engines allow for simulations of a rather broad range of terrains or assets. In addition, most games and simulators allow for battlefield scenarios to be played out at a ``faster then real-time" speed which is a highly desirable feature for the rapid development and testing of data-driven approaches.

% Here we discuss past and current efforts on how games and simulators, together with AI algorithms, have been adapted to simulate military missions, and how these simulations can impact the future battlefield. Further, AI interactions with a future commander equipped with visual augmentation systems are introduced.

% The main contributions of this work:
% \begin{itemize}
%     \item A comprehensive bibliography on how games and simulators have been adapted for use as a platform to develop  AI algorithms for military applications.
%     \item Our current efforts on converting the game \textit{StarCraft II} and the \textit{OpSim} simulator to emulate a combat scenario developed by Military Subject Matter Experts (SMEs), including customized maps, unit skins, and their built-in characteristics, to mimic realistic military scenarios.
%     \item A clear vision for how games and AI algorithms can be used in combination to shape the future battlefield is described in this work. Further, a clear path for transitioning this technology from simulation to the military domain, and the impact of virtual reality and visual augmentation systems is presented.
% \end{itemize}

The paper is organized as follows. In the Background and Related Work section, we offer motivation for research on AI-based tools for Command and Control (C2) decision-making support, and describe a number of prior related efforts to develop approaches to such tools. In particular, we offer arguments in support of using machine learning that can leverage data produced using simulation / wargaming runs, to achieve affordable AI solutions. In the following section, we outline briefly the project that yielded the results described in the paper, and explain how critical for such a research is to find or adapt a game or a simulation with an appropriate set of features. Then, we describe in detail two case studies. In the first, we describe how we adapted a fast and popular gaming system to approximate partly-realistic military operation and how we used it in conjunction with a reinforcement learning algorithm  to train an “artificial commander” (a trained agent) that commands a blue force to fight an opposing enemy force (red force). In the second case study, we describe a similar exploration using a realistic military simulation system. Then we discuss approaches to overcoming common challenges that are likely to arise in transitioning such future technologies to real-world battlefield, and also describe our key findings.

The key contributions of the research described in the paper are two-fold. 
First, we explore and describe two case studies on modeling a partly realistic mission scenario and training an artificial commander leveraging existing (with some adaptations) off-the-shelf deep reinforcement learning algorithms. We show that such trained algorithms can significantly outperform both human and doctrine-based baselines without using any pre-coded expert knowledge, and learn effective behaviors entirely from experience. 
Second, we formulate and empirically confirm a set of features and requirements that a game or simulation engine should meet in order to serve as a viable platform for research and development of reinforcement learning tools for C2. In effect, these are initial recommendations for the researchers who build related experimental and developmental systems. 

% ----------------------------------------------------------------------
%                            RELATED WORK 
% ----------------------------------------------------------------------
\section{Background and Related Work}

% Wargaming is widely used today in the military during training and real world missions. In general, a wargame is considered a type of strategy game that uses rules, data, and procedures to simulate an armed conflict between two or more opposing sides \cite{gortney2010department,caffrey2019wargaming}. According to Burns et al. \cite{burns2015war}, wargaming is a term used to describe ``... a warfare model or simulation whose operation does not involve the activities of actual military forces, and whose sequence of events affects and is, in turn, affected by the decisions made by players representing the opposing sides'' \cite{perla1990art, burns2015war}. This definition of wargaming emphasizes that the purpose of the simulated conflict (i.e., the wargame) is not to simply win, it is to understand how a series of decisions made by adversaries (i.e., the opposing sides) impact one another. In military applications, wargames are often used to study the nature of potential conflicts with real and/or hypothetical adversaries, technologies, and their corresponding capabilities.

We start by defining important terms used throughout this research work. \textit{Wargame} is defined in this work as a largely manual, strategy game that uses rules, data, and procedures to simulate an armed conflict between two or more opposing sides \cite{gortney2010department,burns2015war,caffrey2019wargaming}, which is used to train military officers and plan military courses of action.
This is different from \textit{games}, which is used here as a fully automated, recreational computer application with well-defined rules and score systems that uses a simulation of an armed conflict as a form of entertainment.
Similarly, \textit{simulators} are used here as a hybrid form of wargames and games. Simulators are fully automated computer applications that aim to realistically simulate the outcome of military battles and courses of action. They are not designed as an entertainment platform but as a tool to aid military planning.

Concerning \textit{command and control} (C2), we use the same definition as \textit{command and control warfighting function} defined in the Army Doctrine Publication No.3 - Operations \cite{army2019army} as the ``related tasks and a system that enable commanders to synchronize and converge all elements of combat power".
Similarly, \textit{AI-support to C2} are AI systems developed to aid human commanders by providing additional information or recommendations in the command and control process.

The past few decades have seen a number of ideas and corresponding research towards developing automated or semi-automated tools that could support decision-making in planning and executing military operations.
DARPA’s JFACC program took place in late 1990’s \cite{kott2004advanced} and developed a number concepts and prototypes for agile management of a joint air battle. Most of the approaches considered at that time involved continuous real-time optimization and re-optimization (as situation continually changes) of routes and activities of various air assets. Also in mid-to-late 1990’s, the Army funded the CADET project \cite{kott2002toward} which explored potential utility of classical hierarchical planning, adapted for adversarial environments, for transforming high-level battle sketch into a detailed synchronization matrix – a key product of the doctrinal Military Decision-Making Process (MDMP). In early 2000’s, DARPA initiated the RAID project \cite{ownby2006reading} which explored a number of technologies for anticipating enemy battle plans, as well as dynamically proposing friendly tactical actions. At the time, game-solving algorithms emerged as the most successful among the technological approaches explored. 

The role of multiple domains and their extremely complex interactions – beyond the traditional kinetic fights to include  political, economic and social effects – were explored in late 2000’s in DARPA’s COMPOEX program \cite{kott2007compoex}. This program investigated the use of interconnected simulation sub-models, mostly system-dynamic models, in order to assist senior military and civilian leaders in planning and executing large-scale campaigns in complex operational environments. The importance of non-traditional warfighting domains such as the cyber domain has been recognized and studied in mid-2010’s by a NATO research group \cite{kott2017assessing} that looked into simulation approaches to assessing mission impacts of cyberattacks and highlighted strong non-linear effects of interactions between cyber, human and traditional physical domains. 

All approaches taken in the research efforts mentioned above – and many other similar ones – have major and somewhat common weaknesses. They tend to require a rigid, precise formulation of the problem domain. Once such a formulation is constructed, they tend to produce effective results. However, as soon as a new element needs to be incorporated into the formulation (e.g., a new type of a military asset or a new tactic), a difficult, expensive, manual and long effort is required to “rewire” the problem formulation and to fine-tune the solution mechanism. And the real world endlessly presents a stream of new elements and entities that must be taken into account. 
In rule-based systems of the 1980’s, a system would become un-maintainable as more and more rules (with often unpredictable interactions) had to be added to represent the real-world intricacies of a domain. Similarly, in optimization-based approaches, an endless number of relations between significant variables, and variety of constraints had to be continually and manually be added (a maintenance nightmare) to represent  real-world intricacies of a domain. In game-based approaches, the rules governing  legal moves and effects of moves for each piece would gradually become hopelessly convoluted as more and more realities of a domain had to be manually contrived and added to game formulation.

In short, such approaches are costly in their representation-building and maintenance. Ideally, we would like to see a system that learns its problem formulation and solution algorithm directly from its experiences in a real or simulated world, without any (or with little) manual programming. Machine learning, particularly reinforcement learning, offers that promise.

In contrast, applicability of machine learning to problems of real-world C2 remains a matter of debate and exploration. Walsh et al. \cite{GamesRANDv1, GamesRANDv2} investigated how the military can use the capacity of dealing with large volumes of data and greater decision speed of machine learning algorithms for military planning and command and control.
Walsh et al. \cite{GamesRANDv1} analyzed and rated the characteristics of ten games, such as \textit{Go}, \textit{Bridge}, and \textit{StarCraft II}, and ten command and control processes, such as intelligence preparation of the battlefield, operational assessment, and troop leading procedures. Example of characteristics are the operational tempo, rate of environment change, problem complexity, data availability, stochasticity of action outcomes, and others. 
Their main conclusion related to using games as a platform for military applications was that real-world tasks are very different from many of the games and environments used to develop and demonstrate artificial intelligence systems, which is mostly due to them having fixed and well-defined rules that are regularly exploited by AI agents.

Related to these games, \textit{StarCraft II} (SC2) is a real-time strategy game where players compete for map domination and resources. The players need to manage resources to expand their bases, build more units, upgrade them, and coordinate tens of units at the same time to defeat their opponent.
Vinyals et al. \cite{vinyals2019grandmaster} presented \textit{AlphaStar}, the first AI agent to learn how to play the full game of SC2 at the highest competitive level.
\textit{AlphaStar} played under the same kind of constraints that humans play under professionally approved conditions.
The AI agent was able to achieve this success using a combination of self-play via reinforcement learning, multi-agent learning, and imitation learning using hundreds of thousands of expert replays.
Sun et al. \cite{sun2018tstarbots} proposed the \textit{TStarBot} AI agent, a combination of reinforcement learning and encoded knowledge of the game, becoming the first learning agent to defeat all levels of the built-in rule-based AI in the \textit{StarCraft II} game.
Han et al. \cite{han2020tstarbot} extended that to \textit{TStarBot-X} with a complete analysis of the learned agent trained under a limited scale of computation resources.
Also using limited computational resources, Wang et al. \cite{wang2020scc} presented \textit{StarCraft Commander}, an AI system that also achieves the highest competitive level but using a model with a third of the parameters used by \textit{AlphaStar} and about one-sixth of the training time.

OpenAI et al. \cite{openai2019dota} was the first AI system to defeat the world champions at the Dota 2 esport game, a partially-observable five-against-five real-time strategy game with high dimensionality of observation and action spaces.
In the Dota 2 game, both teams compete for control of strategic locations of the map and to defend their bases at the corners of the map.
Each agent controls a special hero unit with unique abilities that are used when teams engage in conflict against each other and non-player-controlled units.
OpenAI was able to solve this complex task by scaling existing reinforcement learning algorithms, in this case, Proximal Policy Optimization (PPO) \cite{schulman2017proximal} with Generalized Advantage Estimation (GAE) \cite{schulman2015high}, to train using thousand of GPUs over ten months with a specialized neural-network architecture to handle the long time horizons of the game.

Boron and Darken \cite{boron2020developing} investigated the use of deep reinforcement algorithms to solve simulated, perfect information, turn-based, multi-agent, small tactical engagement scenarios based in military doctrine. The grid-world environment emulated an offensive scenario where the homogeneous units controlled by the reinforcement learning agent could move and attack and the defender was static and combat was resolved using both deterministic and stochastic Lanchester combat models \cite{lanchester1916aircraft, washburn2000lanchester, mackay2006lanchester}. One of the main results was how the learned behavior can be controlled to follow different principles of war such as mass or force \cite{corps2011marine} based on the discount factor for the rewards.

Asher et al. \cite{asher2018adapting} proposed the adoption of militarily relevant tactics to a simulated predator-prey pursuit task that allowed for teams of deep reinforcement learning agents to engage in various tactics through capability modifications such as decoys, traps, and camouflage. Further, this research group has introduced methods for measuring coordination \cite{asher2019multi,zaroukian2019algorithmically,asher2020multi} towards a framework for integrating AI agents into mixed Soldier-agent teams \cite{barton2018coordination,barton2018reinforcement}.

With respect to simulators built using existing game engines, Sun et al. \cite{fu20} applied deep reinforcement learning to the command and control of air defense operations. Digital battlefield environment based on \textit{Unreal Engine} was developed for reinforcement learning training process. The training setup consisted of static and random opponent strategies where attack route and formation of units was either fixed or random. For evaluation, win rate, battle damage statistics, and combat details were compared against human experts. Their experimental results showed that deep reinforcement learning agent achieved higher winning rate than the human experts in fixed and random opponent scenarios. Conversely, in this work, the authors are proposing a real-time decision-making and planning system that is designed using reinforcement learning formalism, which can operate in tandem with the commander. 
% With respect to specific simulation platforms used in such research, several systems are worth highlighting. In synergy with other U.S. Army Futures Command-funded research efforts,
\textit{Unity} game engine was used as a base for a multi-agent robot and sensor simulation for military applications\cite{doonan2020aro} and as a quadrotor simulator \cite{song2020flightmare}, while other simulators, such as \textit{Microsoft AirSim} \cite{airsim2017fsr}, use \textit{Unreal Engine} to simulate interaction between multiple robotic agents and the environment, being flexible enough to allow the implementation of military-relevant tasks, as for example, an autonomous landing task where a drone landed on top of a combat vehicle \cite{goecks2019efficiently}.

With respect to military simulators, the AlphaDogfight Trial was a DARPA sponsored competition~\cite{AlphaDogfight-trials} where an AI controled a simulated F-16 fighter jet in an aerial combat against an Air Force pilot flying in a virtual reality simulator. The goal of this program was to automate air-to-air combat. First place winner, Heron Systems, designed an F-16 AI, which outperformed seven other participating companies and defeated an experienced Air Force pilot with a score of 5-0~\cite{AlphaDogfight} in a simulated dogfight. The outcome of this competition demonstrated that an AI can provide precise aircraft maneuvering that may surpass human abilities. In addition, this F-16 AI opened possibilities of human-machine teaming such that human pilots can address high-level strategies while offloading low level, tedious tactical tasks to an AI. 

Schwartz et al. \cite{schwartz2020ai} described a wargaming support tool that used a genetic algorithm to recommend modifications to an existing friendly course of action (COA), a sequence of decisions to be taken in a military scenario, framed as task scheduling problem with multiple tasks and their respective start times. The system integrated user input to the optimization process to constrain which tasks can be modified, minimum and maximum start times, all the AI settings, and monitor all the recommendations that were being simulated by the AI. The authors showed that their system was able to generate expert-level recommendation to friendly courses of action and support the military decision-making process (MDMP) \cite{reese2015military}.

Given these positive results on wargaming, real-time complex strategy and multi-agent recreational applications, and military simulators, extending these algorithms to military command and control applications as a whole is a promising research avenue that we explore in this research work.

% ----------------------------------------------------------------------
%                       GAMES FOR MILITARY APPLICATIONS 
% ----------------------------------------------------------------------
\section{Games and Simulations for Deep Reinforcement Learning of C2}

In this work, we investigate whether deep reinforcement learning algorithms might support future agile and adaptive command and control (C2) of multi-domain forces that would enable the commander and staff to exploit rapidly and effectively fleeting windows of superiority.

To train reinforcement learning agents for a command and control scenario, a fast running simulator with the appropriate interfaces is needed to enable learning algorithms to run for millions of simulation steps, as is often required by state-of-the-art deep reinforcement learning algorithms \cite{espeholt2018impala, kapturowski2018recurrent}.
Much of what motivated us on adapting a game to the study of AI in C2 applications was the difficulty of finding a simulator that had all the required features to develop machine learning algorithms for C2.
These features include, but are not limited to, being able to:
\begin{itemize}
    \item interact with the simulator, also known as the \textit{environment}, via an application programming interface (API), which includes the ability to query environment states and send actions computed by the agent; 
    \item simulate interactions between agent and environment faster than real-time;
    \item parallelize simulations for distributed training of the machine learning agent, ideally over multiple nodes;
    \item randomize environment conditions during training to learn more robust and generalizable reinforcement learning policies; and
    \item emulate realistic combat characteristics such as terrain, mobility, visibility, communications, weapon range, damage, rate of fire, and other factors, in order to mimic real world military scenarios.
\end{itemize}

While we were unable to find a simulator that met all these requirements out of the box, game-based simulators, such as the \textit{StarCraft II Learning Environment} (SC2LE) \cite{vinyals2017starcraft}, were able to be adapted with the purpose of exploring use of applying artificial intelligence algorithms in potential military C2 applications. 
The next sections describe how we adapted SC2LE and the military simulator \textit{OpSim}, respectively, to be used as our main platform to develop an artificial commander for command and control tasks using reinforcement learning.

% ----------------------------------------------------------------------
%                       CASE STUDY: SC2 
% ----------------------------------------------------------------------
\subsection{Case Study: StarCraft II for Military Applications }\label{sec:case_sc2}

As part of this project, we developed a research prototype command and control (C2) simulation and experimentation capability that included simulated battlespaces using the \textit{StarCraft II Learning Environment} (SC2LE) \cite{vinyals2017starcraft} with interfaces to deep reinforcement learning algorithms via \textit{RLlib} \cite{liang2018rllib}, a library that provides scalable software primitives for reinforcement learning on a high performance computing system.

\textit{StarCraft II} is a complex real-time strategy game in which players balance high-level economic decisions with low-level individual control of potentially hundreds of units in order to overpower and defeat an opponent force. \textit{StarCraft II} has a number of difficult challenges for artificial intelligence algorithms that make it a suitable simulation environment for wargaming, command and control, and other military applications. For example, the game has complex state and action spaces, can last tens of thousands of time-steps, can have thousands of actions selected in real time, and can capture uncertainty due to the partial observability or ``fog-of-war''. Further, the game has heterogeneous assets, an inherent control architecture that in some elements resembles military command and control, embedded objectives that have adversarial nature, and a shallow learning curve for implementation/modification compared to more robust simulations.

\textit{DeepMind}’s SC2LE framework \cite{vinyals2017starcraft} exposes \textit{Blizzard Entertainment}'s \textit{StarCraft II} Machine Learning API as a reinforcement learning environment. This tool provides access to \textit{StarCraft II}, its associated map editor, and an interface for RL agents to interact with \textit{StarCraft II}, getting observations and sending actions.

\begin{figure}[htbp]
    \centerline{\includegraphics[width=0.7\columnwidth]{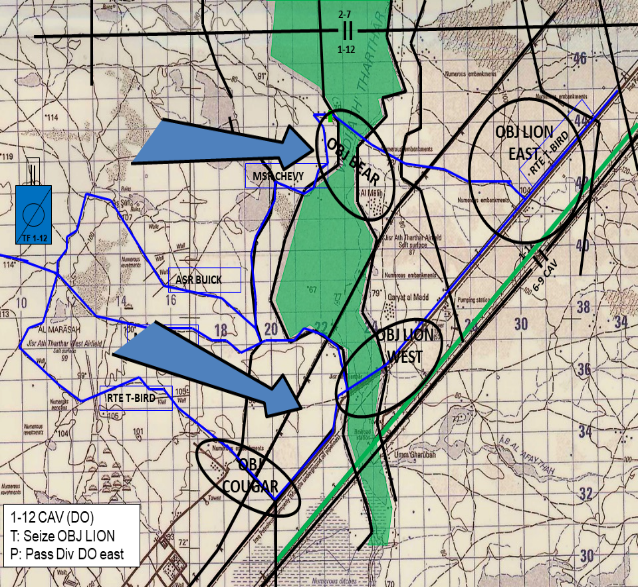}}
    \caption{Illustration of areas of operations in the \textit{TigerClaw} scenario.}
    \label{fig:tigerclaw}
\end{figure}

\begin{figure}[htbp]
    \centerline{\includegraphics[width=0.8\columnwidth]{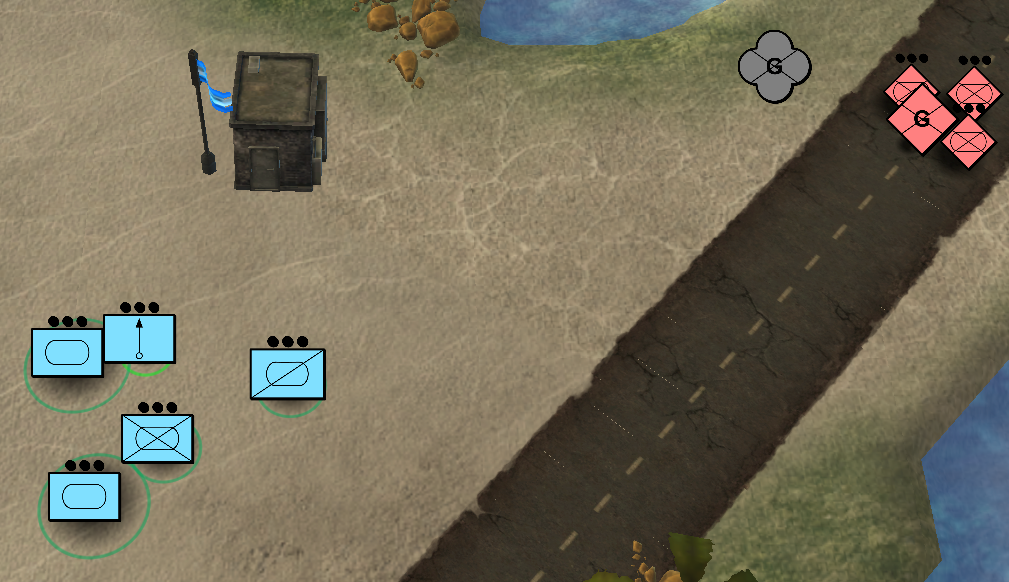}}
    \caption{\textit{StarCraft II} using MILSTD2525 symbols.}
    \label{fig:tigerclaw_sc2_units}
\end{figure}

Using the \textit{StarCraft II Editor} we implemented \textit{TigerClaw}, a Brigade-scale offensive operation scenario \cite{dsifirstyear} to generate a tactical combat scenario within the \textit{StarCraft II} simulation environment, as seen in Fig.~\ref{fig:tigerclaw}.
The game was militarized by re-skining the icons to incorporate MIL-STD-2525C military symbology \cite{std20112525c} and unit parameters (weapons, range, scaling) associated with the \textit{TigerClaw} scenario, as seen in Fig.~\ref{fig:tigerclaw_sc2_units}. 

In \textit{TigerClaw}, the Blue Force's goal is to cross the wadi terrain, neutralize the Red Force, and control certain geographic locations. These objectives are encoded in the game score for use by  reinforcement learning agents and serving as a benchmarking baseline for comparison across different neural network architectures and reward driving attributes. 
The following sections describe the process we used to adapt the map, units, and rewards to this military scenario.

\subsubsection{Map Development}

\begin{figure}[htbp]
    \centerline{\includegraphics[width=0.95\columnwidth]{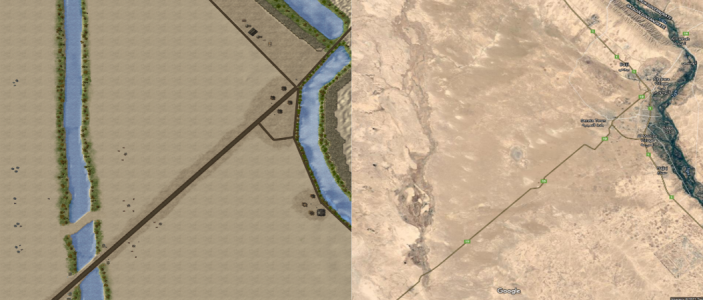}}
    \caption{Modified \textit{StarCraft II}, left panel, map to resemble real world area of operation, right panel.}
    \label{fig:tigerclaw_sc2_map}
\end{figure}

We created a new \textit{Melee Map} for \textit{TigerClaw} scenario using the \textit{StarCraft II Editor}. The map size was the largest available, 256 by 256 tiles, using the \textit{StarCraft II} coordinate system. A wasteland tile set was used as the default surface of the map since it visually resembled a desert region in the area of operations in \textit{TigerClaw}, as seen in Fig.~\ref{fig:tigerclaw_sc2_map}.
After the initial setup, we used the \textit{Terrain} tools to modify the map to loosely approximate the area of operation. The key terrain feature was the impassable wadi with limited crossing points.

Distance scaling was an important factor for the scenario creation. In the initial map, we used the known distance between landmarks to translate \textit{StarCraft II} distance, using its internal coordinate system, into kilometers and latitude and longitude. This translation is important for adjusting weapons range during unit modification and to ensure compatibility with other internal visualization tools that expect geographic coordinates as input.

\subsubsection{Playable Units Modification}

To simulate the \textit{TigerClaw} scenario, we selected \textit{StarCraft II} ``fantastic" units that could be made to approximate the capabilities of realistic military units, even if crudely. The \textit{StarCraft II} units were first duplicated and their attributes modified in the Editor to support the scenario. 
First, we modified the appearance of the units and replaced it with an appropriate MIL-STD-2525C symbol, as seen in Table \ref{tab:tigerclaw_unit_mapping}.

\begin{table}[]
\centering
\caption{Mapping of \textit{TigerClaw} units to \textit{StarCraft II} units.}
\label{tab:tigerclaw_unit_mapping}
\begin{tabular}{@{}cc@{}}
\toprule
\textbf{\textit{TigerClaw} Unit} & \textbf{\textit{StarCraft II} Unit} \\ \midrule
Armor & Siege Tank (tank mode) \\
Mechanized Infantry & Hellion \\
Mortar & Marauder \\
Aviation & Banshee \\
Artillery & Siege Tank (siege mode) \\
Anti-Armor & Reaper \\
Infantry & Marine \\ \bottomrule
\end{tabular}
\end{table}

Other attributes modified for the scenario included, weapon range, weapon damage, unit speed, and unit health (how much damage it can sustain). Weapon ranges were discerned from open source materials and scaled to the map dimensions. Unit speed was established in the \textit{TigerClaw} operations order and fixed at that value. The attributes for damage and health were estimated, with the guiding principle of maintaining a conflict that challenges both sides. Each \textit{StarCraft II} unit usually had only one weapon making it challenging to simulate the variety of armaments available to a company size unit. Additional effort to increase the accuracy of unit modifications will require wargaming subject matter experts.

Additionally, the Blue Force units were modified so that they would not engage offensively or defensively unless specifically commanded by the player or learning agent in control. To control the Red Forces, we used two different strategies. The first strategy was to include a scripted course of action, a set of high-level actions to take, for Red Force movements that is executed in every simulation. The units default aggressiveness attributes controlled how it engaged Blue. The second strategy was to let a \textit{StarCraft II} bot AI control the Red Force to execute an all-out attack, or suicide as it is termed in the \textit{Editor}. The built-in \textit{StarCraft II} bot has several difficulty levels (1–10) which dictate the proficiency of the bot. The bot levels indicate their proficiency where a level 1 is a fairly rudimentary bot that can be easily defeated and level 10 is a very sophisticated bot that uses information not available to players (i.e., a cheating bot). Finally, environmental factors such as fog-of-war were toggled across experiments to investigate their impact. This was done in the following manner:

\subsubsection{Game Score and Reward Implementation}

Reward function is an important component of reinforcement learning and it controls how the agent reacts to environmental changes by giving them positive or negative reward for each situation. We incorporated the reward function for the \textit{TigerClaw} scenario in \textit{StarCraft II} and our implementation overrode the internal game scoring system. The default scoring system in \textit{StarCraft II} rewarded players for the resource value of their units and structures. Our new scoring system focused on gaining and occupying new territory as well as destroying the enemy. 

\begin{figure}[htbp]
    \centerline{\includegraphics[width=0.75\columnwidth]{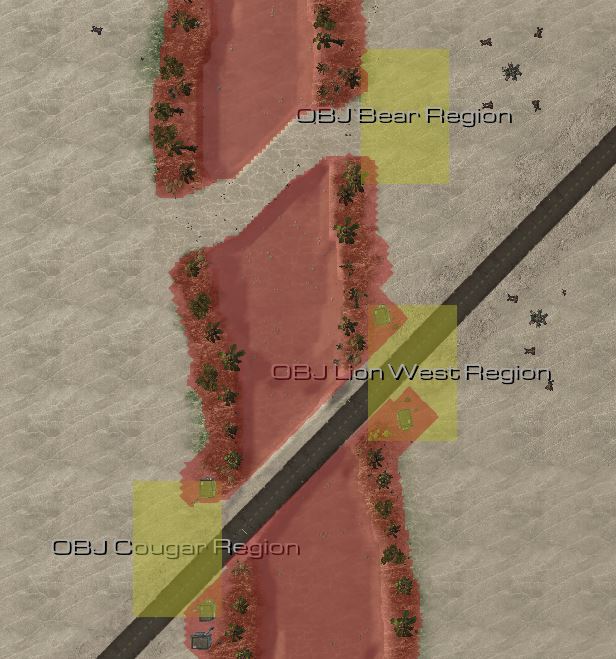}}
    \caption{Regions and objectives used as part of the reward function in the custom \textit{TigerClaw} scenario.}
    \label{fig:tigerclaw_sc2_map_reward}
\end{figure}

Our reward function awarded $+10$ points for the Blue Force crossing the wadi (river) and $-10$ points for retreating back. In addition, we awarded $+10$ points for destroying a Red Force unit and $-10$ points if a Blue Force unit was destroyed.
In order to implement the reward function, it was necessary to first use the \textit{StarCraft II} map editor to define the various regions and objectives of the map. Regions are areas, defined by the user, which are utilized by internal map triggers that compute the game score, as seen in Fig.~\ref{fig:tigerclaw_sc2_map_reward}.

Additional reward components can be also integrated based on the the Commander’s Intent in the \textit{TigerClaw}, or other scenario, warning orders. Ideally, the reward function will attempt to train the agent to create optimal behavior that is perceived as reasonable by a military subject matter expert. 

\subsubsection{Deep Reinforcement Learning Results}

\begin{figure}[htbp]
    \centerline{\includegraphics[width=0.95\columnwidth]{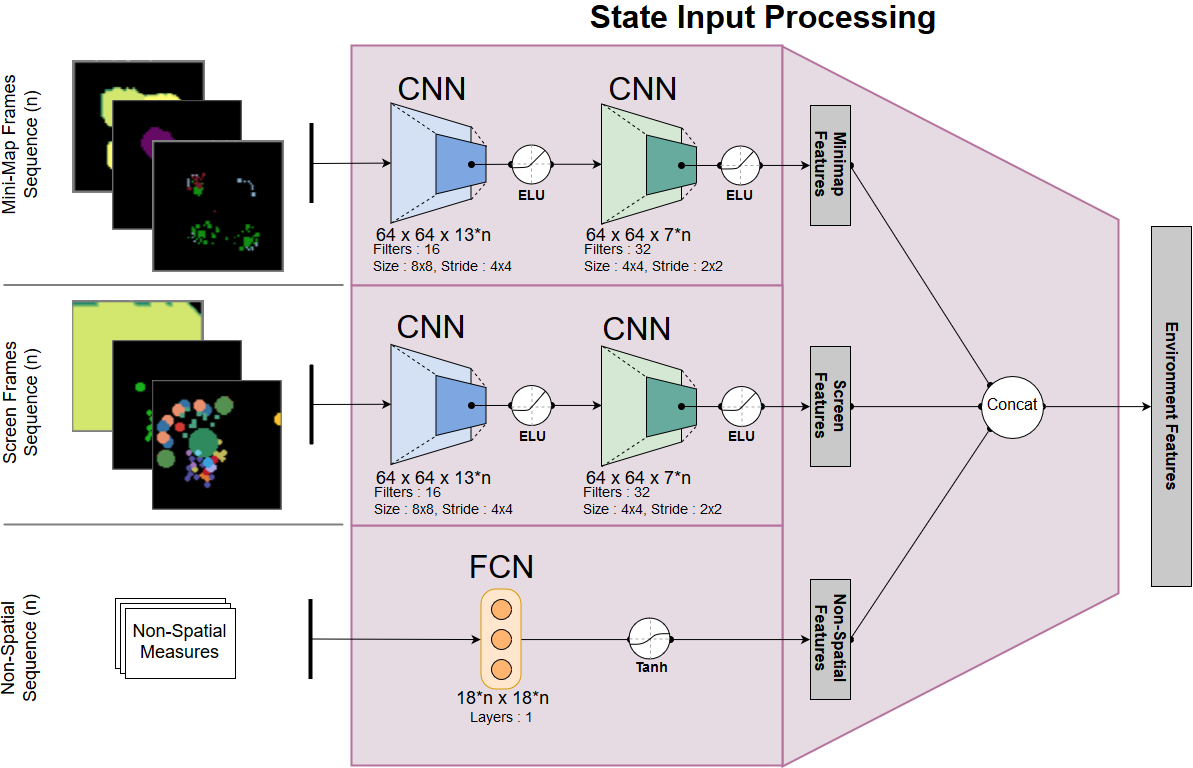}}
    \caption{Input processing for the custom \textit{TigerClaw} scenario.}
    \label{fig:sc2_input}
\end{figure}

Using the custom \textit{TigerClaw} map, units, and reward function, we trained a multi-input and multi-output deep reinforcement learning agent adapted from Waytowich et al 2019 \cite{waytowich2019ICML}. The RL agent was trained using the Asynchronous Advantage Actor Critic (A3C) algorithm  \cite{mnih2016asynchronous}.
In this tactical version of the \textit{StarCraft II} mini-game, as shown in Fig.~\ref{fig:sc2_input}, the state-space consists of 7 mini-map feature layers of size 64x64 and 13 screen feature layer maps of size 64x64 for a total of 20 64x64 2D images. Additionally, it also consists of 13 non-spatial features containing information such as player resources and build queues.
The mini-map and screen features were processed by identical 2-layer convolutional neural networks (top two rows) in order to extract visual feature representations of the global and local states of the map, respectively. The non-spatial features were processed through a fully-connected layer with a non-linear activation. These three outputs were then concatenated to form the full state-space representation for the agent.

The actions in \textit{StarCraft II} are compound actions in the form of functions that require arguments and specifications about where that action is intended to take place on the screen. For example, an action such as ``attack'' is represented as a function that would require the $x-y$ attack locations on the screen. The action space consists of the action identifier (i.e., which action to run), and two spatial actions ($x$ and $y$) that are represented as two vectors of length $64$ real-valued entries between $0$ and $1$. 

\begin{figure}[htbp]
    \centerline{\includegraphics[width=0.95\columnwidth]{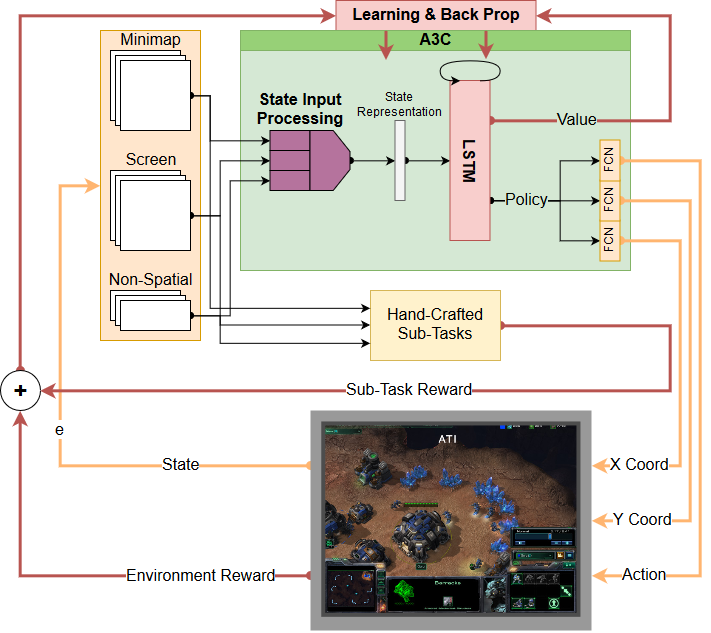}}
    \caption{Schematic diagram of the full A3C reinforcement learning agent and its connection to the \textit{StarCraft II} environment representing the  \textit{TigerClaw} scenario.}
    \label{fig:sc2_architecture}
\end{figure}

The architecture of the A3C agent we use is similar to the Atari-net agent \cite{mnih2015human}, which is an A3C agent adapted from Atari to operate on the \textit{StarCraft II} state and actions space. We make one slight modification to this agent and add a long short-term memory (LSTM) layer \cite{hochreiter1997long}, which adds memory to the model and improves performance \cite{mnih2016asynchronous}. The complete architecture of our A3C agent is shown in Fig.~\ref{fig:sc2_architecture}.

\begin{figure}[htbp]
    \centerline{\includegraphics[width=0.95\columnwidth]{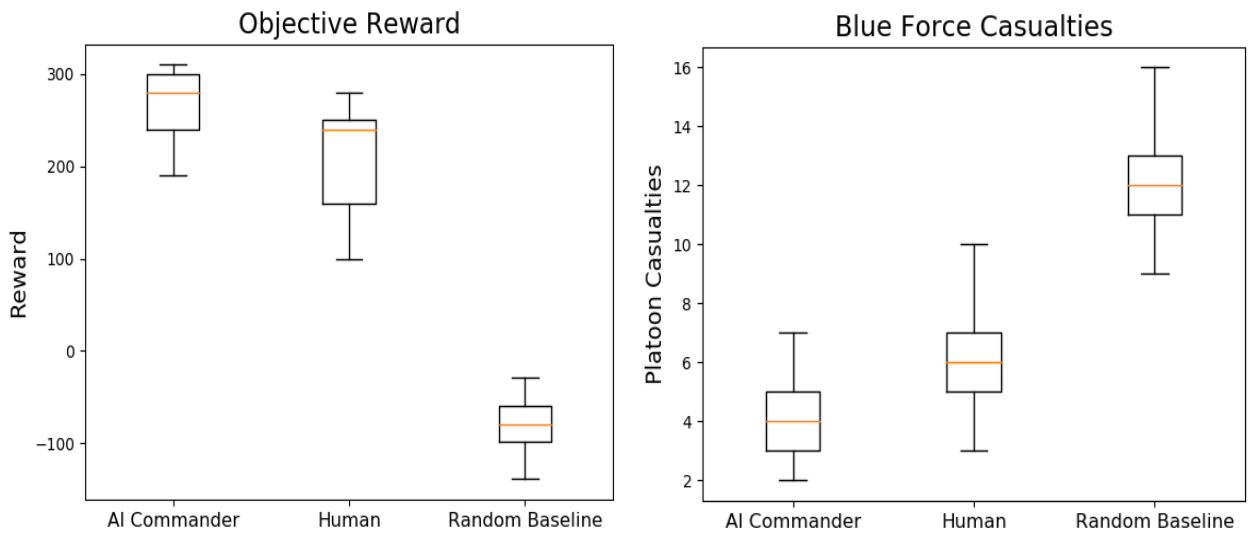}}
    \caption{Total reward and Blue Force casualties of the trained AI commander (A3C agent) compared to human and random agent baselines. The AI commander is able to achieve a reward that is comparable (and slightly better) than the human baseline while taking a reduced number of Blue Dorce casualties.}
    \label{fig:sc2_a3c_results}
\end{figure}

The A3C models were trained with 20 parallel actor-learners using 8,000 simulated battles against a built-in \textit{StarCraft II} bot operating on hand crafted rules.
Each trained model was tested on 100 rollouts of the agent on the \textit{TigerClaw} scenario. The models were compared against a random baseline with randomized actions as well as a human player playing 10 simulated battles against the \textit{StarCraft II} bot.
Fig.\ref{fig:sc2_a3c_results} show plots of total episode reward and number of Blue Force casualties during the evaluation rollouts. We see that the AI commander has not only achieved comparable performance compared to a human player, but has also performed slightly better at the task, while also reducing Blue Force casualties.

% ----------------------------------------------------------------------
%                       CASE STUDY: OPSIM 
% ----------------------------------------------------------------------
\subsection{Case Study: Reinforcement Learning using OpSim: a Military Simulator}\label{sec:case_opsim}

The \textit{TigerClaw} scenario, as described in the previous case study, was also implemented in the \textit{OpSim} \cite{surdu1999opsim} simulator.
\textit{OpSim} is a decision support tool developed by Cole Engineering Services Inc. (CESI) that provides planning support, mission execution monitoring, mission rehearsal, embedded training, and mission execution monitoring and re-planning. \textit{OpSim} integrates with \textit{SitaWare C4I} command and control, a critical component of Command Post Computing Environment\endnote{Command Post Computing Environment (CPCE) webpage: https://peoc3t.army.mil/mc/cpce.php} (CPCE) fielded by \textit{PEO C3T} allowing all levels of command to have shared situational awareness and coordinate operational actions, thus making it an embedded simulation that connects directly to operational mission command.

\textit{OpSim} is fundamentally constructed as an extensible Service Oriented Architecture (SOA) based simulation and is capable of running faster than current state of the art simulation environments such as One Semi-Automated Forces\endnote{One Semi-Automated Forces (OneSAF) webpage: https://www.peostri.army.mil/OneSAF} (OneSAF) \cite{wittman2001onesaf, parsons2005onesaf}, MAGTF Tactical Warfare Simulation\endnote{MAGTF Tactical Warfare Simulation (MTWS) webpage: https://coleengineering.com/capabilities/mtws} (MTWS) \cite{blais1994marine}. \textit{OpSim} is designed to run much faster that wall clock time, and can run 30 replications of the \textit{TigerClaw} mission, which would take two hundred and forty hours if run serially in real time. Output of a simulation plan in \textit{OpSim} include an overall ranking of Blue Force plans based on criteria such as ammunition expenditure, casualties, equipment loss, fuel usage, and others. The \textit{OpSim} tool however was not originally designed for AI applications and had to be adapted by incorporating interfaces to run reinforcement learning algorithms.

An \textit{OpenAI Gym} \cite{brockman2016openai} interface was developed to expose simulation state and offer simulation control to external agents with the ability to supply actions for select entities within the simulation, as well as the amount of time to simulate before responding back to the interface.
The observation space consists of 17 features vector where the observation space is partially observable based on each entities' equipment sensors. Unlike the \textit{StarCraft II} environment, our \textit{OpSim}-based prototype of an artificial C2 agent currently does not use image inputs or spatial features from the screen images. The action space primarily consists of simple movements and engagement attacks, as shown below:
\begin{itemize}
    \item \textbf{Observation space: }damage state, $x$ location, $y$ location, equipment loss, weapon range, sensor range, fuel consumed, ammunition consumed, ammunition total, equipment category, maximum speed, perceived opposition entities, goal distance, goal direction, fire support, taking fire, engaging targets.
    \item \textbf{Action space: }no operation, move forward, move backward, move right, move left,  speed up,  slow down, orient to goal, halt, fire weapon, call for fire, react to contact.
    \item \textbf{Reward function: }friendly damaged ($-0.5$), friendly destroyed ($-1.0$), enemy damaged ($0.5$), enemy destroyed ($1.0$), $-0.01*km$ from goal destination at every step.
\end{itemize}

\subsubsection{Experimental Results}

Two types of ``artificial commanders" were developed for the \textit{OpSim} environment. The first is based on an expert designed rule engine provided as part of \textit{OpSim} and developed by Military Subject Matter Experts (SMEs) using military doctrinal rules. The second is a reinforcement learning-based long short-term memory (LSTM) \cite{hochreiter1997long} deep neural network with a multi-input and multi-output trained with Advantage Actor Critic (A2C) algorithm \cite{mnih2016asynchronous}. \textit{OpSim}'s custom \textit{Gym} interface supports multi-agent training where each force can use an either rule or learning-based commander.

\begin{figure}[htbp]
    \centerline{\includegraphics[width=0.95\columnwidth]{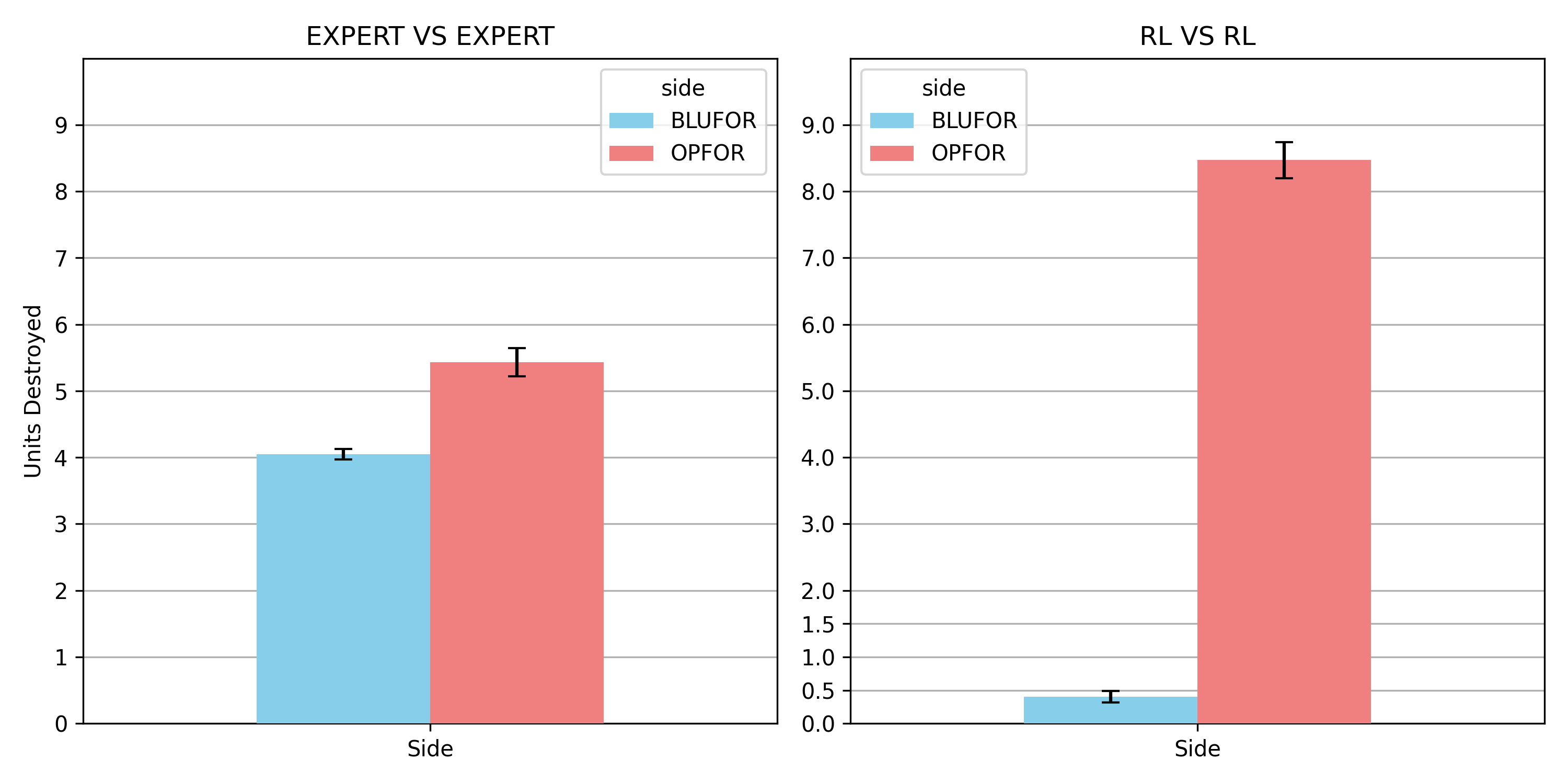}}
    \caption{Unit casualties comparison between rule-based (expert) and reinforcement learning-based (RL) commanders in the \textit{TigerClaw} scenario.}
    \label{fig:opsim_results}
\end{figure}

\begin{figure}[htbp]
    \centerline{\includegraphics[width=0.95\columnwidth]{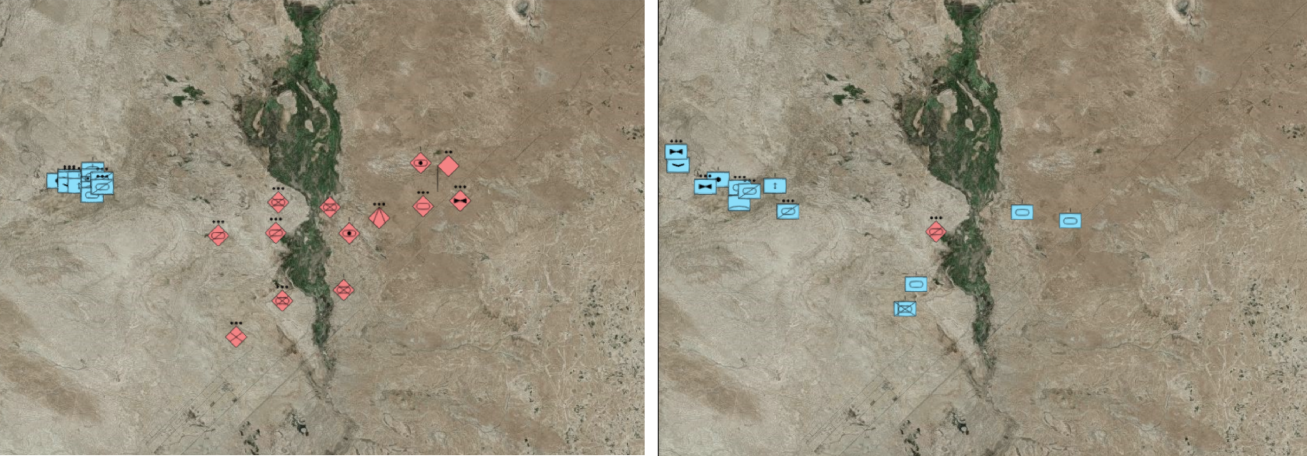}}
    \caption{Beginning (left) and end (right) of a reinforcement learning-based commander rollout in \textit{OpSim} illustrating a learned strategy by the Blue Force to engage only with the most lethal units while protecting vulnerable assets.}
    \label{fig:opsim_rollout}
\end{figure}

The policy network was trained on a high performance computing facility with 39 parallel workers collecting 212,000 simulated battles in 44 hours. 
The trained models were evaluated with 100 rollout simulation results using the frozen policy at a checkpoint with the highest mean reward for the Blue Force, in this case, a rolling average of $195$ for the Blue Force policy mean reward and $-317$ for the Red Force policy mean reward.
Analysis of 100 rollout, as seen in Figure \ref{fig:opsim_results}, show that the reinforcement learning-based commander minimizes Blue Force casualties from about 4 to 0.4, on average, when compared to the rule-based commander and increases Red Force casualties from 5.4 to 8.4, on average, in the same comparison. This outcome is reached by employing a strategy to engage using only combat armor companies and fighting infantry companies. The reinforcement learning-based commander has learned a strategy to utilize Blue Force's (BLUFOR) most lethal units with Abrams and Bradleys vehicles while protecting vulnerable assets from engaging with the Red Force (OPFOR), as seen in a snapshot of the beginning and end of one rollout shown in Figure \ref{fig:opsim_rollout}. 

% ----------------------------------------------------------------------
%                         FUTURE OF THE FIELD 
% ----------------------------------------------------------------------
\section{Challenges and Future Applications with Virtual Reality}

Transition to practice is a common and major challenge. An approach that might reduce challenges of transitioning AI-based C2 tools – as we are beginning to explore in the cases studies described above -- is the adaptation of games and simulators with built-in AI capabilities to enable course of action (COA) development and wargaming.
This path provides the commander and staff with opportunities to rapidly explore in great detail multiple friendly courses of action (COAs) against multiple enemy COAs, capability or asset allocation and employment, terrain or environmental changes, and various visualizations to better understand how an adversary might respond to selected COAs in a simulated battle. Therefore, the inclusion of AI enabled games into current military practices might accelerate, supplement, or otherwise enhance current practices in COA and wargaming development and decisions.

Another challenge is operator interfaces that often fail to gain end-user acceptance. To this end, another capability potentially adapted from the gaming industry are head-mounted displays (HMD) which afford users the ability to interact with content in virtual reality (VR), augmented reality (AR), and mixed reality (MR) settings - collectively XR. Traditionally, these technologies have been used in industry to provide immersive experiences for storytelling, to enable remote real-time collaboration in shared virtual spaces, and to augment natural perception with holographic overlays. For real-world military applications, XR technologies can be used to provide a more intuitive display of multi-domain information, enable distributed collaboration, and to enhance situational awareness and understanding. While the Army has already invested in XR technologies for training, through the Synthetic Training Environment\endnote{Synthetic Training Environment (STE) webpage: https://asc.army.mil/web/portfolio-item/synthetic-training-environment-ste/}, and to enhance lethality, through the Integrated Visual Augmentation System (IVAS)\endnote{Integrated Visual Augmentation System (IVAS) webpage: https://www.peosoldier.army.mil/Program-Offices/Project-Manager-Integrated-Visual-Augmentation-System/}, it has relied heavily on advances made in the gaming industry to do so. In fact, many military XR projects are developed using either Unity or Unreal - the two most popular game development engines.

For this project, initial integration has begun to pull live data from the \textit{StarCraft II} simulation into a networked XR environment \cite{dennison2020accelerated}. The goal is to allow multiple, decentralized users with the ability to not only examine AI-generated courses of action in an immersive platform, but to interact with the AI and other human decision-makers collaboratively in real time. Prior research \citep{10.3389/frobt.2019.00082} has shown that understanding of spatial information, like that depicted in a three-dimensional war-game, can be enhanced by visualization in an HMD. Furthermore, rendering of the tactical information in a game engine like Unity enables researchers with the ability to precisely control all aspects of how the battlefield is visualized and related user interfaces. Interoperability between this immersive XR interface and other military program of record command and control information systems, such as CPCE discussed in the \textit{OpSim} case study, is also being investigated to allow explicit comparison between immersive and non-immersive tools.

% ----------------------------------------------------------------------
%                         FINDINGS AND DISCUSSION 
% ----------------------------------------------------------------------
\section{Findings and Discussion}

Our research highlights several important aspects of developing a training system for AI algorithms in military-relevant games and battlefield simulators for command and control.

First, with respect to both the adapted game and battlefield simulator, we found that an important aspect is \textit{simulation speed}.
Reinforcement learning algorithms are notorious for requiring millions, or even billions \cite{espeholt2018impala, kapturowski2018recurrent}, of data samples from the environment to train the best-performing policies, so fast simulation cycles are essential to achieve results in a reasonable time.
In our \textit{StarCraft II} experiments we were able to simulate 8 million training steps, or 12,800 battles, per day with 35 CPU workers, and that was still not fast enough given that our agents required over 40 million training steps to achieve satisfactory results.
Given that the simulations run at least for days, another important aspect for a simulator to be considered for this type of research work is \textit{scalability}. Advances in computing capacity has propelled the field of reinforcement learning in the past decade and thus a simulator for AI research needs to support distributed computing at scale.
%If the available simulator does not support continuous execution of a training run over the course of several days, the researcher should investigate alternatives to reliably restore training agent checkpoints.
Yet another aspect is \textit{adaptability}. The simulator should give the experiment designer the flexibility to model the desired military task, including terrain, assets, and fog-of-war.

Second, with respect to training reinforcement learning agents in simulators, due to the nature of the exploratory behavior of these reward-driven algorithms, sometimes learning agents are able to exploit loopholes in simulator dynamics and learn unrealistic behaviors, usually unintended by the simulator designer, in order to maximize the received reward \cite{amodei2016concrete}.
This type of exploitation is often not detected during training, and are discovered only when carefully investigating the trained policy behavior after the training session. 
For example, in our early experiments, we found that the agent was able to exploit a simulator deficiency allowing armored assets to cross impassable terrain.  In this sense, reinforcement learning can be a tool to detect and strengthen simulators. 

Third, with respect to the training procedure, it should comprise of  diverse training data, for instance terrain, assets, and location, so the learning artificial commander is able to develop a more robust and general policy and be able to address variability that exists in reality. How diverse and how much variability these scenarios should contain is still an open research question.
During training, it also helps to evaluate the learning agent periodically, either against a rule or doctrine-based commander or against another learning artificial commander.
Once training is complete and the agent is evaluated in test cases, we found that visualization tools are essential to uncover if the artificial commander learned any unintended behavior although more visualization tools are needed to understand each detail of the commander's actions and plans.

Finally, we found that reinforcement learning has been able, at least in a set of cases we explored, to outperform both human and doctrine-based baselines without access to prior human knowledge of the task, which becomes even more relevant when task complexity increases and human-dictated rules are more difficult to specify.
However, there are still open questions on how human-like the strategies learned are, or how human-like they need to be if that is the case.
Another concern is that the reinforcement learned policies are willing to sacrifice assets in the battle if that leads to a higher reward value at the end of the scenario, which may not be an acceptable decision for a human commander.
We address this issue when handcrafting the reward function by penalizing the agent for assets lost, however, given that the reward function is also composed of additional goals, this solution is non-trivial to balance in practice \cite{amodei2016concrete}.

% ----------------------------------------------------------------------
%                            CONCLUSIONS 
% ----------------------------------------------------------------------
\section{Conclusions}

As artificial intelligence (AI) algorithms have been successfully learning how to solve complex games, their impact on military applications is probably imminent.
We envision the need to develop agile and adaptive AI support tools for the future battlefield and multi-domain operation scenarios, under the main assumption that the future flow of information and speed of operation will likely exceed the capabilities of the current human staff if the command and control processes remain largely manual.
This includes leveraging AI algorithms to analyze battlefield information from multiple sources, from both red and blue forces, and correctly identify and exploit emerging windows of superiority.

In this research work, we explore two case studies on modeling a partially realistic mission scenario, and training artificial commanders that leverage existing (with some adaptations) off-the-shelf deep reinforcement learning algorithms. We show that such trained algorithms can outperform significantly both human and doctrine-based baselines without using any pre-coded expert knowledge, and learning effective behaviors entirely from experience. 
Furthermore, we formulate and empirically confirm a set of features and requirements that a game or simulation engine must meet in order to serve as a viable platform for research and development of reinforcement learning tools for C2.

% ----------------------------------------------------------------------
%                            ACKNOWLEDGEMENT 
% ----------------------------------------------------------------------
\begin{acks}

Portions of the Related Work, \textit{StarCraft II} and \textit{OpSim} case studies also appear in the DEVCOM Army Research Laboratory report ARL-TR-9192 \cite{dsifirstyear}.
\end{acks}

% ----------------------------------------------------------------------
%                             FUNDING 
% ----------------------------------------------------------------------
\section{Funding}

\textbf{The author(s) disclosed receipt of the following financial support for the research, authorship, and/or publication of this article:} Research was sponsored by the Army Research Laboratory and was accomplished partly under Cooperative Agreement [W911NF-20-2-0114]. The views and conclusions contained in this document are those of the authors and should not be interpreted as representing the official policies, either expressed or implied, of the Army Research Laboratory or the U.S. Government. The U.S. Government is authorized to reproduce and distribute reprints for Government purposes notwithstanding any copyright notation herein.

% ----------------------------------------------------------------------
%                            REFERENCES 
% ----------------------------------------------------------------------
\theendnotes

\bibliographystyle{SageV}
\bibliography{references}

\end{document}